\documentclass{article}

\usepackage{arxiv}

\usepackage[utf8]{inputenc} 
\usepackage[T1]{fontenc}    
\usepackage[colorlinks=true, citecolor=blue, linkcolor=blue, urlcolor=blue]{hyperref}
\usepackage{url}            
\usepackage{booktabs}       
\usepackage{siunitx}
\usepackage{amsfonts}       
\usepackage{bm}
\usepackage{mathtools}
\usepackage{amsmath}
\usepackage{amssymb}
\usepackage{multirow}
\usepackage{nicefrac}       
\usepackage{microtype}      
\usepackage{lipsum}		
\usepackage{graphicx}
\usepackage{doi}
\usepackage[acronym]{glossaries}
\glsdisablehyper
\usepackage[font={scriptsize,sf},labelfont=bf]{caption}
\usepackage[backend=biber, style=ieee, hyperref=true, natbib=false]{biblatex}

\newacronym{IF}{IF}{integrate-and-fire}
\newacronym{LIF}{LIF}{leaky integrate-and-fire}
\newacronym{ANN}{ANN}{artificial neural network}
\newacronym{CNN}{CNN}{convolutional neural network}
\newacronym{PPO}{PPO}{proximal policy optimization}
\newacronym{BPTT}{BPTT}{backpropagation through time}
\newacronym{MFCC}{MFCC}{mel-frequency cepstrum coefficient}
\newacronym{MSE}{MSE}{mean squared error}
\newacronym{MSD}{MSD}{mean squared difference}

\newcommand{\vm}[1]{\bm{#1}}

\title{Memory-enriched computation and learning in spiking neural networks through Hebbian plasticity}

\author{Thomas Limbacher$^a$ \And Ozan \"Ozdenizci$^{a,b}$ \And Robert Legenstein$^a$\thanks{To whom correspondence should be addressed. E-mail: robert.legenstein@igi.tugraz.at}}

\date{
	$^a$Institute of Theoretical Computer Science, Graz University of Technology, 8010 Graz, Austria \\
	$^b$TU Graz - SAL Dependable Embedded Systems Lab, Silicon Austria Labs, 8010 Graz, Austria \\[1ex]
	\texttt{\{thomas.limbacher, ozan.ozdenizci, robert.legenstein\}@igi.tugraz.at}\\[4ex]%
	\today
}

\renewcommand{\headeright}{}
\renewcommand{\undertitle}{}
\renewcommand{\shorttitle}{Memory-enriched computation and learning in spiking neural networks through Hebbian plasticity}

\hypersetup{
pdftitle={Memory-enriched computation and learning in spiking neural networks through Hebbian plasticity},
pdfsubject={cs.NE,cs.LG},
pdfauthor={Thomas Limbacher, Ozan \"Ozdenizci, Robert Legenstein},
pdfkeywords={Spiking Neural Networks, Hebbian Plasticity, Memory, Few-shot learning},
}

\begin{document}
\maketitle

\begin{abstract}
Memory is a key component of biological neural systems that enables the retention of information over a huge range of temporal scales, ranging from hundreds of milliseconds up to years. While Hebbian plasticity is believed to play a pivotal role in biological memory, it has so far been analyzed mostly in the context of pattern completion and unsupervised learning. Here, we propose that Hebbian plasticity is fundamental for computations in biological neural systems. We introduce a novel spiking neural network architecture that is enriched by Hebbian synaptic plasticity. We show that Hebbian enrichment renders spiking neural networks surprisingly versatile in terms of their computational as well as learning capabilities. It improves their abilities for out-of-distribution generalization, one-shot learning, cross-modal generative association, language processing, and reward-based learning. As spiking neural networks are the basis for energy-efficient neuromorphic hardware, this also suggests that powerful cognitive neuromorphic systems can be build based on this principle.
\end{abstract}


\section{Introduction}
Spiking Neural Networks (SNNs) are a well-established model of neural computation \cite{maass2001pulsed}. In contrast to conventional \glspl{ANN}, neurons in an SNN communicate via stereotypical pulses---so-called spikes---and temporally integrate incoming information in their membrane potential. Since these are key features of biological neurons, SNNs are heavily used to model information processing in the brain. Furthermore, SNNs are well-suited for implementation in neuromorphic hardware, leading to highly energy-efficient AI applications. 

For a long time, SNNs have been inferior to \glspl{ANN} in terms of performance on standard pattern recognition tasks. However, a number of recent advances in SNN research have changed the picture, showing that SNNs can achieve performances similar to \glspl{ANN} \cite{tavanaei2019deep}. In particular, the use of surrogate gradients for SNN training \cite{bellec2018long,huh2018gradient,zenke2018superspike,neftci2019surrogate} and the use of longer adaptation time constants in recurrent SNNs have been instrumental in this respect \cite{salaj2021spike}. Nevertheless, SNNs still lack many capabilities of their biological counterparts---for some of which biologically implausible \gls{ANN} solutions have been proposed.  

Since computations in SNNs have---in contrast to computations in feed-forward \glspl{ANN}---a strong temporal component, they have been proposed to be particularly suited for temporal computing tasks \cite{salaj2021spike}. Here, it has turned out that the ability to retain information on several time scales is crucial. The most basic time-constant in spiking neurons is the membrane time constant on the order of tens of milliseconds. In principle, arbitrary time constants can be realized by recurrent connections in recurrent SNNs. However, such recurrent retention of information is rather brittle and hard to learn. Instead, there were several suggestions to utilize longer time constants available in biological neuronal circuits such as short-term plasticity \cite{maass2002real,mongillo2008synaptic} on the order of 100s of milliseconds, and adaptation time constants of neurons \cite{bellec2018long,salaj2021spike} on the order of seconds. The inclusion of such time constants has been shown to extend the computational capabilities of SNNs. However, typical cognitive tasks are frequently situated on a much slower time scale of minutes or longer. For example, when we watch a movie, we have to rapidly memorize facts in order to follow the story and draw conclusions as the narrative evolves. For such tasks, time constants on the order of seconds are insufficient. 

Here we consider Hebbian synaptic plasticity \cite{hebb2005organization} as a mechanism to extend the range of time constants and therefore the computational capabilities of SNNs. Hebbian synaptic plasticity is abundant in both neocortex and hippocampus \cite{bliss1973long,malenka1999long,rioult2000learning}. While many forms, in particular in sensory cortical areas, are believed to shape processing on a very slow developmental scale, there is also evidence for rapid plasticity that can in principle be utilized for online processing on the behavioral time scale, most prominently in the hippocampus \cite{bittner2017behavioral,zhao2022rapid}. We show that Hebbian synaptic plasticity renders SNNs rather flexible in terms of their computational as well as learning capabilities. This single principle can give rise to SNNs that are capable of out-of-distribution generalization, one-shot learning, cross-modal generative association, answering questions about stories of various types, and learning to play card games from rewards. Hence, our results show that Hebbian plasticity enhances the computational capabilities of SNNs in several directions. This suggests that Hebbian plasticity is a central component of information processing in the brain which is tightly interwoven with cognitive neuronal processing. Since local Hebbian plasticity can easily be implemented in neuromorphic hardware, this also suggests that powerful cognitive neuromorphic systems can be build based on this principle.

\section{Results}
\label{sec:results}

\subsection{Spiking Neural Networks with associative memory}
\label{sec:results-model}
We consider networks of standard \gls{LIF} neurons modeled in discrete time steps $\Delta t$. The membrane potential $V_j$ of neuron $j$ at time $t$ is given by
\begin{equation}
\label{eq:Vm}
    V_j(t + \Delta t) = \alpha V_j(t) + (1 - \alpha) I_j(t) - \vartheta z_j(t),
\end{equation}
where $\alpha$ defines the membrane potential decay per time step. The total synaptic input current $I_j(t)= \sum_i W_{ji} z_i(t)$ is given by the sum over all pre-synaptic neuron's output spike trains weighted by the corresponding synaptic weights $W_{ji}$. When the neuron's membrane potential $V_j(t)$ is above some threshold $\vartheta$, the neuron spikes ($z_j(t)=1$) and the membrane potential is reset (last term in \eqref{eq:Vm}). If the neuron does not spike, we define $z_j(t)=0$.

\begin{figure}
\centering
\includegraphics{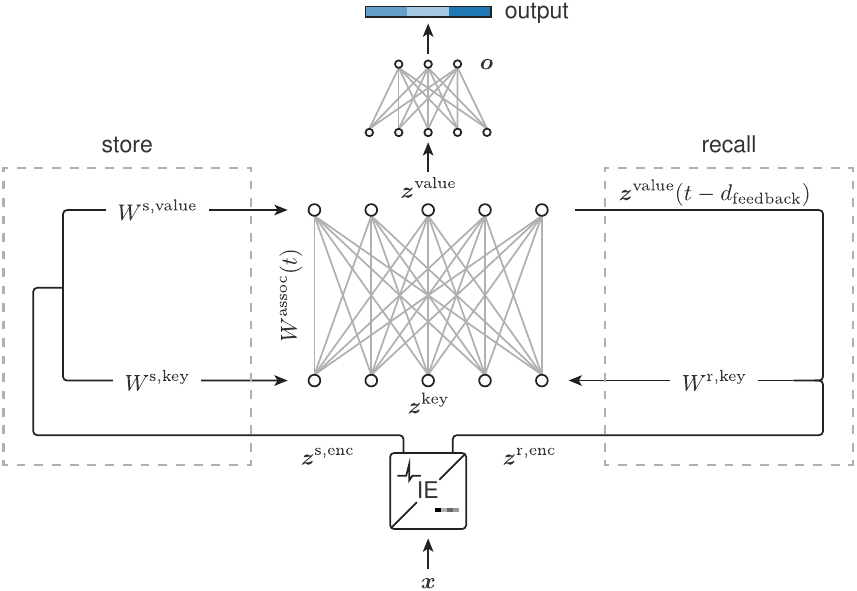}
\caption{Schematic of the SNN model. Inputs $\vm{x}$ are encoded by an input encoder (IE). Memory induction: The encoded input $\vm{z}^\mathrm{s,enc}$ activates some neurons in the key- and value-layer through weight matrices $W^\mathrm{s,key}$ and $W^\mathrm{s,value}$, respectively. Since the key-neurons $\vm{z}^\mathrm{key}$ and the value-neurons $\vm{z}^\mathrm{value}$ are pre- and post-synaptic to the synapses in $W^\mathrm{assoc}$, their activity induces weight changes there. Memory recall: The encoded input $\vm{z}^\mathrm{r,enc}$ activates some neurons in the key-layer through the weight matrix $W^\mathrm{r,key}$. This activity activates some neurons in the value-layer through synapses $W^\mathrm{assoc}$, thus potentially recalling some information that has been stored previously. Finally, value neurons $\vm{z}^\mathrm{value}$ project to a layer of output neurons $\vm{o}$. To allow for a memory recall based on previously recalled information, activity in the value-layer is fed back to the key-layer with some delay $d_\mathrm{feedback}$.
}
\label{fig:fig1}
\end{figure}


The considered network model is shown in Fig.~\ref{fig:fig1}. At the core of the network is a heteroassociative memory \cite{willshaw1969non}, that is, a single layer feed-forward SNN. Here, spiking neurons $\vm{z}^{\mathrm{key}}$ in the {\em key-layer} project to neurons $\vm{z}^{\mathrm{value}}$ in the {\em value-layer} with synaptic weights $W_{ji}^{\mathrm{assoc}}$ that are subject to rapid Hebbian plasticity. We used \num{100} neurons in the key- and value-layer respectively in all simulations. The use of this simple hetero-associative memory architecture is motivated from the hippocampal circuitry where it was shown that rapid plasticity is found in the connections from region CA3 to CA1, resembling a similar single-layer architecture with key-layer corresponding to CA3 and the value-layer corresponding to CA1 \cite{bittner2017behavioral,zhao2022rapid}. From a machine learning perspective, this architecture can be motivated from memory-augmented neural networks, a class of \gls{ANN} models that were shown to outperform standard \glspl{ANN} in memory-dependent computational tasks. This class includes networks with key-value memory systems such as memory-networks \cite{weston2015towards,sukhbaatar2015end}, Hebbian memory networks \cite{limbacher2020hmem}, and transformers \cite{vaswani2017attention}. The latter have been shown to be particularly powerful for language-processing, giving rise to language models such as GPT-3 \cite{brown2020language}.

In our model, neurons in the key-layer (key neurons) receive input from two spiking neuron populations, $\vm{z}^{\mathrm{s,enc}}$ and $\vm{z}^{\mathrm{r,enc}}$, 
responsible for storing (s) and recalling (r) information to and from the memory respectively (see below).
Similarly, neurons in the value-layer (value neurons) receive input from two sites, namely from $\vm{z}^{\mathrm{s,enc}}$ and from the key layer neurons $\vm{z}^{\mathrm{key}}$.
Consider an input $\vm{x}$ from which some aspects should be stored in memory. It is first encoded by an input encoder (IE in Fig.~\ref{fig:fig1}; the architecture of the encoder depends on the task at hand, see below) giving rise to a spike response $\vm{z}^{\mathrm{s,enc}}$. The synaptic connections to the key-layer activate some key neurons, while the synaptic connections to the value-layer activate some value neurons. As these neurons are pre- and post-synaptic to the synapses in  $W^{\mathrm{assoc}}$, this induces weight changes there (see \emph{Methods} for the Hebbian plasticity model). Which information about the input $\vm{x}$ is stored depends on the synaptic weights from $\vm{z}^{\mathrm{s,enc}}$ to the key- and value-layers.

Now consider an input $\vm{x}$ that should trigger a memory recall. In this case, the encoded input $\vm{z}^{\mathrm{r,enc}}$ activates the key neurons through the weight matrix $W^{\mathrm{r,key}}$ giving rise to activity in the key-layer. This activity activates neurons in the value-layer through association synapses $W^{\mathrm{assoc}}$, thus potentially recalling information that has been stored previously. Finally, these neurons project to a layer of output neurons $\vm{o}$. For some tasks, it might be beneficial to perform a recall based on previously recalled information. For example, when you are asked "Can Tweety fly?", you may first recall that Tweety is a canary and then remember that canaries can fly to arrive at the correct answer. We therefore included a feedback loop in the model from the value neurons to the key neurons (see loop on the right side of Fig.~\ref{fig:fig1}). In this way, recalled activity can influence the recall itself after some delay. We set the feedback delay $d_\mathrm{feedback}$ to \SI{1}{ms} if not otherwise stated.

\subsection{Memorizing associations}
\label{sec:results-random_associations}
We first tested the ability of our model to one-shot memorize associations and to use these associations later when needed. Here we conducted experiments on a task that requires to form associations between random continuous-valued vectors and integer labels that were sequentially presented to the network.

\begin{figure}
\centering
\includegraphics{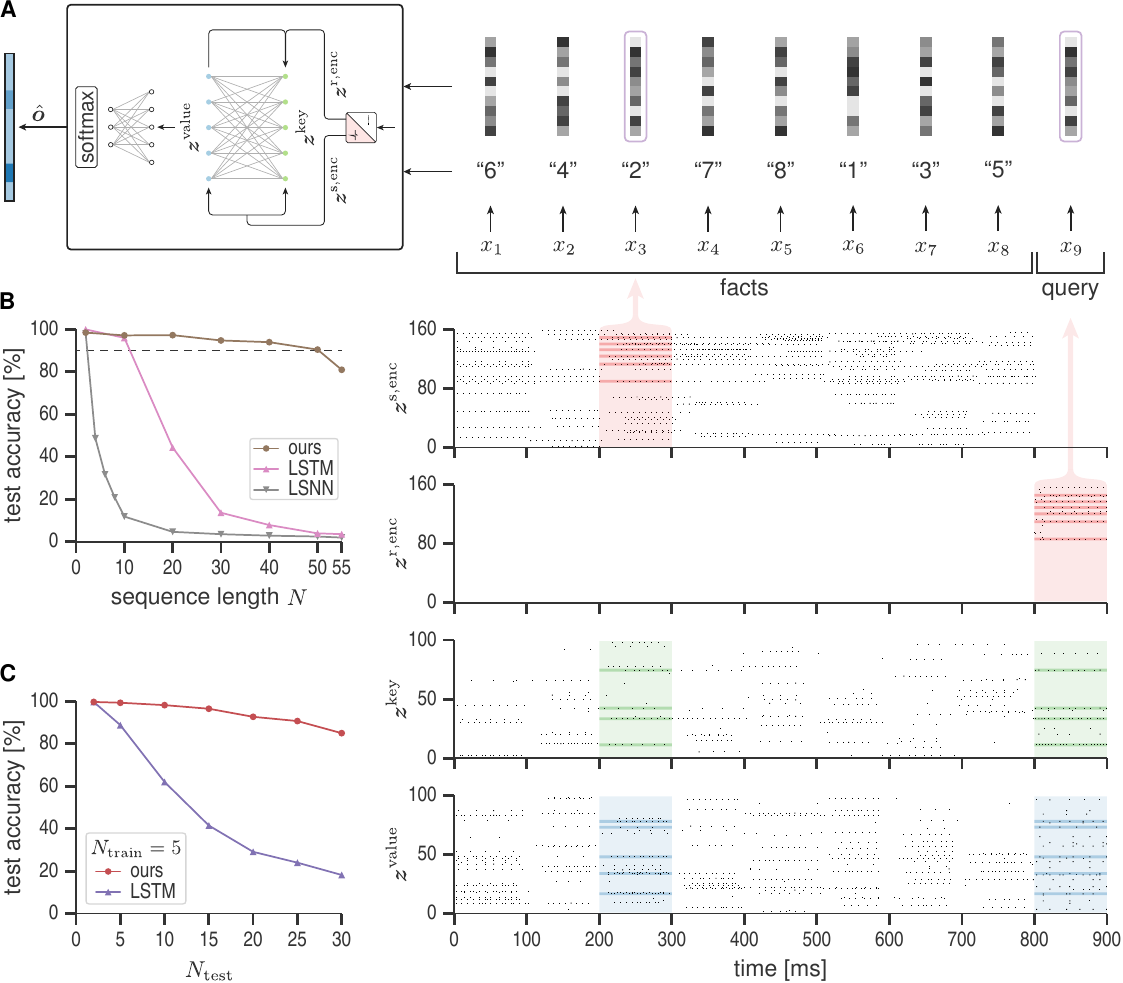}
\caption{Association task and out-of-distribution generalization.
\textbf{(A)} Schematic of the network model (left), the network input for an input sequence of length $N=8$ (top right), and the network activity after training (bottom right).
Eight \num{10}-dimensional random vectors along with an associated label from \num{1} to \num{8} are presented sequentially as facts ($x_1$ to $x_8$). After all those vector-label pairs have been presented to the network, it receives a query vector ($x_9$). The query vector is equal to one of the vectors presented as facts (randomly selected; vector with label \num{2} in the shown example). The network is required to output the label of the query vector. Inputs are encoded as spike trains $\vm{z}^\mathrm{s,enc}$ and $\vm{z}^\mathrm{r,enc}$ (\SI{100}{ms} for each input). Synaptic connections to the key-layer activate some key neurons $\vm{z}^{\mathrm{key}}$, while the synaptic connections to the value-layer activate some value neurons $\vm{z}^{\mathrm{value}}$. Spikes in these layers during storing of the third vector-label pair $x_3$ and during recall are shown within red, green, and blue rectangles, respectively. Neurons that are both active during storing of $x_3$ and recalling are highlighted with saturated color.
\textbf{(B)} Performance comparison of our model with an LSTM network and an LSNN in this task. Shown is the test accuracy for various sequence lengths $N$. While the accuracy of our model stays above \SI{90}{\percent} for sequences with up to \num{50} vector-label pairs, the performance of the LSTM network and the LSNN quickly decreases with increasing sequence length.
\textbf{(C)} Out-of-distribution generalization capability. We trained models with a sequence length of $N_\mathrm{train}=5$ and evaluated the models generalization capability to test sets with shorter and longer sequence lengths $N_\mathrm{test}$. Comparison to LSTM network: While both models generalize to new test sets with shorter sequences, our model shows superior generalization to longer sequences.
}
\label{fig:fig2}
\end{figure}

In each instance of this task, we randomly drew $N$ real vectors where each element of a vector was sampled from a uniform distribution on the interval $[0, 1)$. We then generated input sequences of $N$ tuples each containing one of the random vectors and an associated label from \num{1} to $N$ (Fig.~\ref{fig:fig2}A, right). After all those vector-label pairs have been presented to the network, it received a query vector. The query vector was equal to one of the $N$ vectors and was randomly selected for each input sequence. The network was required to output the label of the query vector.

The model was trained for \num{4250} iterations using \gls{BPTT} \cite{werbos1990backpropagation,bellec2018long}. We used two dense layers as input encoder in this task. Each layer consisted of \num{80} \gls{LIF} neurons. One layer was used to encode each of the random input vectors and another layer was used to encode the integer labels. Inputs were applied to these layers for \SI{100}{\ms} each, giving rise to spike trains $\vm{z}^\mathrm{s,enc}$ and $\vm{z}^\mathrm{r,enc}$ (see Section \emph{Model and training details} in the Supplementary for more details to the model and the training setup). Fig.~\ref{fig:fig2}A (bottom right) shows the network activity after training for one test example with a sequence length $N$ of eight vector-label pairs.

In Fig.~\ref{fig:fig2}B we compare the performance of our model for various sequence lengths (number of vector-label pairs) to the standard generic artificial and spiking recurrent network models: the standard LSTM network \cite{hochreiter1997long} and the long short-term memory spiking neural network (LSNN) \cite{bellec2018long,salaj2021spike}. The LSTM network consisted of \num{100} LSTM units. The LSNN consisted of \num{150} regular spiking and \num{150} adaptive neurons. For both models, we used the same architecture for the input encoder and the output layer as in our model. For short sequences, the performance of the LSTM network and the LSNN is comparable to our model. While the accuracy of the LSTM and LSNN drastically drops at some point, the accuracy of our model stays above \SI{90}{\percent} for sequences containing up to \num{50} vector-label pairs.

To test the out-of-distribution generalization capability of our model in this task, we trained it with a sequence length $N_\mathrm{train}$ of five vector-label pairs and tested the model on shorter and longer sequences of up to \num{30}. Note that the labels in this task were randomly chosen from $\{1, \dots, 30\}$ and, consequently, the output layer was of size \num{30}. In Fig.~\ref{fig:fig2}C, we compare its performance to an LSTM network in terms of the test accuracy. While both models generalize to shorter sequences, our model shows superior generalization to longer sequences.


\subsection*{One-shot Learning}
\label{sec:results-one_shot_learning}
While standard deep learning approaches need large numbers of training examples during training, humans can learn new concepts based on a single exposure (one-shot learning) \cite{lake2017building}. A large number of few-shot learning approaches have been proposed using artificial neural networks \cite{snell2017prototypical,finn2017model}, but biologically plausible SNN models are very rare \cite{scherr2020one}. We wondered whether Hebbian plasticity could endow SNNs with one-shot learning capabilities.   
To that end, we applied our model to the problem of one-shot classification on the Omniglot \cite{lake2015human} data set. Omniglot consists of \num{1623} handwritten characters from \num{50} different alphabets. There are \num{20} examples of each character, each hand drawn by a different person. Following \cite{vinyals2016matching}, we resized all the image to $28 \times 28$ and augmented the data set with rotations in multiples of \num{90} degrees. We trained on \num{1028} characters (\num{4112} classes in total with rotations) and tested on \num{423} characters (\num{1692} classes with rotations).

\begin{figure}
\centering
\includegraphics{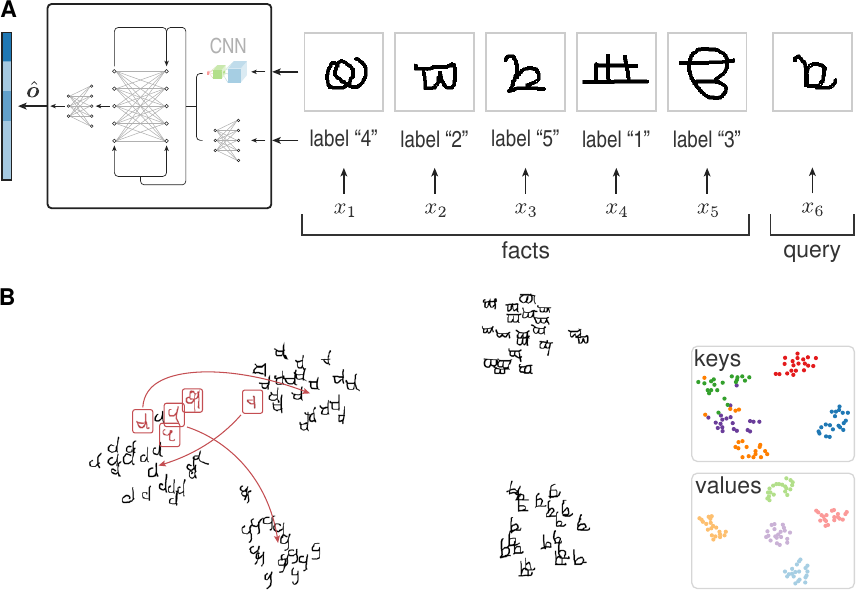}
\caption{Omniglot one-shot task and a visualization of the embeddings learned by the encoder \gls{CNN} in this task.
\textbf{(A)} Schematic of the network model (left) and the network input (right). Five images from the Omniglot data set along with an associated label from \num{1} to \num{5} are presented sequentially as facts ($x_1$ to $x_5$). After all those image-label pairs have been presented to the network, it receives a query image ($x_6$). The network is required to output the label that appeared together with an image of the same class as the query image.
\textbf{(B)} A t-SNE visualization of the embeddings learned by the encoder \gls{CNN}. A subset of the Tengwar script is shown (an alphabet in the test set). Misclassified characters are highlighted in red along with arrows pointing to the correct cluster. Inset: A t-SNE visualization of the learned key- and value-representation of the inputs. Colors indicate character class.}
\label{fig:fig3}
\end{figure}

In each instance of this task, we randomly drew five different Omniglot classes. We then generated input sequence of tuples each containing a randomly drawn instance of one of the five Omniglot classes and an associated label from \num{1} to \num{5} (Fig.~\ref{fig:fig3}A right). After all those image-label pairs have been presented to the network, it received a query image. The query image showed another randomly drawn sample from one of these five Omniglot classes. The network was required to output the label that appeared together with an image of the same class as the query image (\num{1}-shot \num{5}-way classification).

We trained our model for \num{200} epochs with \num{200} iterations per epoch. In each iteration we randomly drew a batch of \num{256} input sequences, each containing five different Omniglot classes along with an associated label from \num{1} to \num{5}, and one query image. We used a \gls{CNN} as input encoder for the Omniglot image, which was pre-trained using the prototypical loss \cite{snell2017prototypical} and then converted into a spiking \gls{CNN} by using a threshold-balancing algorithm \cite{diehl2015fast,sengupta2019going} (see Section \emph{Converting pre-trained CNNs} in the Supplementary for details to the conversion algorithm). The spiking \gls{CNN} was then fine-tuned during end-to-end training. We treated the grayscale values of an Omniglot image as a constant input current, applied for \SI{100}{\ms} to \num{784} \gls{LIF} neurons, to produce input spikes to the \gls{CNN}. The final layer of the \gls{CNN} consisted of \num{64} \gls{LIF} neurons. A single dense layer consisting also of \num{64} \gls{LIF} neurons was used to encode the integer labels as spike trains with a duration of \SI{100}{\ms} per label (see Section \emph{Model and training details} in the Supplementary for more details to the model and the training setup).

The rationale for this network architecture is that, given a suitable generalizing representation of the character, the Hebbian weight matrix can easily associate characters to labels, thus performing one-shot memorization of previously unseen classes to arbitrary labels. In biology, suitable representations could emerge from evolutionary optimized networks potentially fine-tuned by unsupervised plasticity processes. In our setup, these representations are provided by the \gls{CNN} encoder, where the prototypical loss ensures that similar inputs are mapped to similar representations \cite{snell2017prototypical}. The network model can thus be seen as a spiking implementation of a prototypical network. However, the biologically unrealistic nearest-neighbor algorithm used to determine the output of the latter is replaced here by a simple hetero-associative memory. 
In Fig.~\ref{fig:fig3}B we show a sample t-SNE visualization of the embeddings produced by the spiking \gls{CNN} that was used as image encoder in this task. Despite the shown characters being rather diverse, the network is able to represent them as well-separated clusters. This clustering can also be observed in the learned key- and value-representations of the inputs (inset of Fig.~\ref{fig:fig3}B). Overall, the SNN achieved an accuracy of \SI{92.2}{\percent} when tested on 1-shot 5-way classification of novel character classes.

\subsection{Cross-modal associations}
\label{sec:results-audio_image}
Humans are able to imagine features of previously encountered stimuli. For example, when you hear the name of a person, you can imagine a mental image of its face. Here, in contrast to the associations considered above, not just classes are associated but (approximate) mental images. We therefore asked whether Hebbian plasticity can enable SNNs to perform such cross-modal associations. We trained our model in an autoencoder-like fashion. We used the FSDD \cite{fsd2016} and the MNIST \cite{lecun2010mnist} data set in this task. FSDD is an audio/speech data set consisting of recordings of spoken digits. The data set contains \num{3000} recordings of \num{6} speakers (\num{50} of each digit per speaker).

\begin{figure}
\centering
\includegraphics{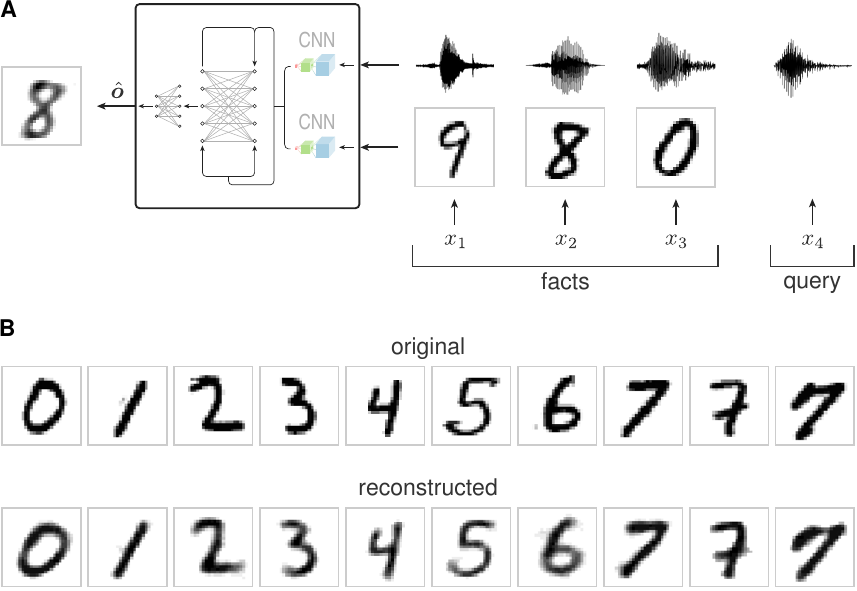}
\caption{Audio to image synthesis task and examples of generated images.
\textbf{(A)} Schematic of the network model (left) and the network input (right). Three audio-image pairs are presented sequentially as facts. Input pairs ($x_1$, $x_2$, and $x_3$) contain a spoken digit from the FSDD data set and an image of the same digit from the MNIST data set. After all those audio-image pairs have been presented to the network, it receives an audio query ($x_4$). The network is required to generate an image of the handwritten digit that appeared together with the spoken digit of the same class as the audio query.
\textbf{(B)} Example MNIST images from the test set (top) and the corresponding images that were reconstructed by the network (bottom). The reconstructed images are not just typical images for the digit classes, but rather images that are very similar to the images presented previously with the audio cues (compare the three rightmost image pairs).}
\label{fig:fig4}
\end{figure}

In each instance of the task we randomly drew three unique digits between \num{0} and \num{9}. For each digit we then generated a tuple containing a randomly drawn instance of this digit as audio file from the FSDD data set, and a randomly drawn image of the same digit from the MNIST data set (Fig.~\ref{fig:fig4}A right). After these audio-image pairs have been presented to the network, it received an additional audio query. The audio query was another randomly drawn instance from the FSDD data set of one of the previously presented digits. The network was required to generate the image of the handwritten digit that appeared together with the spoken digit of the same class as the audio query.

We used one \gls{CNN} to encode the audio input (more specifically the \glspl{MFCC}, see Section \emph{Model and training details} in the Supplementary), and one \gls{CNN} to encode the MNIST images. The \glspl{CNN} were pre-trained on FSDD/MNIST classification tasks respectively and the pre-trained models were converted into spiking \glspl{CNN} by using the threshold-balancing algorithm \cite{diehl2015fast,sengupta2019going} (see Section \emph{Converting pre-trained CNNs} in the Supplementary for details to the conversion algorithm). We removed the final classification layers of the \glspl{CNN} and used the penultimate layer consisting of \num{64} \gls{LIF} neurons as the encoding of the input stimuli. The spiking \gls{CNN} was then fine-tuned during end-to-end training (see Section \emph{Model and training details} in the Supplementary for more details to the model and the training setup). Following the value layer of the model, the image reconstruction was produced by a two-layer fully-connected network with \num{256} and \num{784} \gls{LIF} neurons, respectively.

In Fig.~\ref{fig:fig4}B we show example MNIST images from the test set and the corresponding images that were reconstructed by the network. One can see that not just a typical image for the digit class was imagined by the network, but rather an image that is very similar to the image presented previously with the audio cue. This shows that the network did not just memorize the digit class in its associative memory, but rather features that benefit the reconstruction of this specific sample.
To quantify the reconstruction performance of our model, we computed the \gls{MSD} between the image produced by the network and all MNIST images in an input sequence. The \gls{MSD} was $0.03\pm0.02$ (mean $\pm$ standard deviation; median was \num{0.02} with a lower and upper quartile of \num{0.01} and \num{0.03}, respectively) between the reconstructed image and the target image, and $0.1\pm 0.04$ between the reconstructed image and the two other MNIST images in the input sequence (statistics are over \num{1000} examples in the test set). 

\subsection{Question Answering}
\label{sec:results-babi}
The bAbI data set \cite{weston2015towards} is a standard benchmark for cognitive architectures with memory. The data set contains \num{20} different types of synthetic question-answering (QA) tasks, designed to test a variety of reasoning abilities on stories. Each of these tasks consist of a sequence of sentences followed by a question whose answer is typically a single word (in a few tasks, answers are multiple words; see Table S1 in the Supplementary for example stories and questions). We provided the answer to the model as supervision during training, and it had to predict it at test time on a separate test set. The performance of the model was measured using the average error on the test data set over all tasks and the number of failed tasks (according to the convention of \cite{weston2015towards}, a model had failed to solve a task if the test error was above \SI{5}{\percent} for that task).

Each instance of a task consists of a sequence of $M$ sentences $\langle x_m, \dots, x_M\rangle$, where the last sentence is a question, and an answer $a$. We represent each word $j$ in a given sentence $x_m$ by a one-hot vector $\vm{w}_{m,j}$ of length $V$ (where $V$ is the vocabulary size). We limited the number of sentences in a story to \num{50} (similar to previous work \cite{sukhbaatar2015end,henaff2017tracking}).

We used a dense layer consisting of \num{80} \gls{LIF} neurons as input encoder in this task. We found it helpful to let the model choose for itself which type of sentence encoding to use. We therefore used a learned encoding (see Section \emph{Model and training details} in the Supplementary and \cite{henaff2017tracking}). Each sentence was encoded as a spike train with a duration of \SI{100}{ms}. Similar to previous work \cite{sukhbaatar2015end,henaff2017tracking}, we performed three independent runs with different random initializations and report the results of the model with the highest validation accuracy in these runs.

In Table \ref{tab:tab1} we compare our model to the Spiking RelNet \cite{plank2021long} and to the H-Mem model \cite{limbacher2020hmem}, a non-spiking memory network model. Similar to the feedback-loop from the value-layer to the input of the key-layer in our model, the H-Mem model can utilize several memory accesses conditioned on previous memory recalls. The results of H-Mem were reported for a single memory access (1-hop) and three memory accesses (3-hop). It turned out that multiple memory hops were necessary to solve some of the bAbI tasks. Similarly, we found that in our model, an instantaneous feedback loop (with a delay $d_\mathrm{feedback}$ of \SI{1}{\ms}) struggled with a few tasks that could be solved with a delay of \SI{30}{\ms}, corresponding roughly to three hops during the network inference that lasted \SI{100}{\ms}. We compared the performance of these models in terms of their mean error, error on individual tasks, and the number of failed tasks. 
Our model with \SI{1}{\ms} feedback solved \num{13} of the \num{20} tasks. By increasing the feedback delay $d_\mathrm{feedback}$ from \SI{1}{\ms} to \SI{30}{\ms} it was able to solve \num{16} of the \num{20} tasks. This result indicates that the spiking network can make use of multiple memory accesses in an asynchronous manner, i.e., simply through a delayed feedback loop without the need for discrete memory access steps.  
The spiking RelNet model solved \num{17} of the \num{20}. Note however that this model is much more complex, makes heavy use of weight sharing and employs pre-trained LSNNs for word embeddings.

\begin{table}
\centering
\caption{Test error rates (in \si{\percent}) on the \num{20} bAbI QA tasks. Comparison of our model to the Spiking RelNet model \cite{plank2021long} and to the non-spiking memory-based model H-Mem \cite{limbacher2020hmem} performing one memory access (1-hop) and three memory accesses (3-hop). Shown are mean error, error on individual tasks, and the number of failed tasks (according to the convention of \cite{weston2015towards}, a model had failed to solve a task if the test error was above \SI{5}{\percent} for that task; results of the alternative models were taken from the respective papers). Keys: mem.~acc.~= number of memory accesses; fb.~delay~= synaptic delay $d_\mathrm{feedback}$ in feedback loop.}
\sffamily\scriptsize\selectfont
\begin{tabular}{@{}lrrrrr@{}}
\toprule
&\multirow{3}{*}{\shortstack{Spiking\\RelNet}}&\multicolumn{2}{c}{H-Mem}&\multicolumn{2}{c}{ours}\\
\cmidrule(lr){3-4} \cmidrule(l){5-6}
&&\multicolumn{2}{c}{mem.~acc.}&\multicolumn{2}{c}{fb.~delay}\\
Task && 1-hop & 3-hop & \SI{1}{ms} & \SI{30}{ms}\\
\midrule
1: Single Supporting Fact           & 1.0 & 0.0  & 0.0  & 0.0  & 0.0  \\
2: Two Supporting Facts             & -   & 64.2 & 0.2  & 10.4 & 3.3  \\ 
3: Three Supporting Facts           & -   & 58.6 & 26.9 & 58.6 & 59.2 \\
4: Two Arg. Relations               & 0.1 & 0.0  & 0.0  & 0.0  & 0.0  \\
5: Three Arg. Relations             & 2.3 & 4.1  & 1.3  & 1.4  & 2.1  \\
6: Yes/No Questions                 & 0.4 & 12.2 & 1.2  & 28.2 & 4.2  \\
7: Counting                         & 1.4 & 0.8  & 0.8  & 3.2  & 3.4  \\
8: Lists/Sets                       & 0.9 & 0.4  & 0.5  & 0.8  & 0.5  \\
9: Simple Negation                  & 0.7 & 15.5 & 3.3  & 1.4  & 2.7  \\
10: Indefinite Knowledge            & 1.7 & 21.3 & 1.5  & 5.8  & 3.9  \\
11: Basic Coreference               & 2.1 & 0.1  & 0.0  & 0.0  & 0.0  \\
12: Conjunction                     & 4.2 & 0.0  & 0.0  & 0.0  & 0.1  \\
13: Compound Coref.                 & 3.6 & 2.3  & 0.0  & 0.3  & 0.0  \\
14: Time Reasoning                  & 0.0 & 7.9  & 1.1  & 4.4  & 2.7  \\
15: Basic Deduction                 & 0.0 & 1.0  & 0.0  & 0.0  & 0.0  \\
16: Basic Induction                 & -   & 54.2 & 54.8 & 54.1 & 53.4 \\
17: Positional Reasoning            & 2.3 & 38.8 & 28.7 & 37.7 & 15.7 \\
18: Size Reasoning                  & 0.2 & 4.8  & 1.9  & 0.8  & 1.8  \\
19: Path Finding                    & 3.7 & 74.7 & 77.1 & 66.5 & 65.8 \\
20: Agent's Motivations             & 0.4 & 0.0  & 0.0  & 0.0  & 0.0  \\
\cmidrule(lr){2-2} \cmidrule(lr){3-4} \cmidrule(l){5-6}
Mean error                          & 1.5 & 18.0 & 10.0 & 13.9 & 10.9 \\
Failed tasks (err. $>\SI{5}{\percent}$) & 3 & 9 & 4    & 7 & 4 \\
\bottomrule
\end{tabular}
\label{tab:tab1}
\end{table}

\subsection{Reinforcement Learning}
\label{sec:results-reinforcement_learning}
While supervisory signals are arguably scarce in nature, it is well-established that animals learn from rewards \cite{schultz2002getting}. In order to test whether memory can also serve SNNs in the reward-based setting, we evaluated our model on an episodic reinforcement learning task. The task is based on the popular children's game \emph{Concentration}. The game \emph{Concentration} requires good memorization skills and is played with a deck of $n$ pairs of cards. The cards in each pair are identical. At the start of each game the cards are shuffled and laid out face down. A player's turn consists of flipping over any two of the cards. If the cards match then they are removed from the game and the current player moves again. Otherwise the cards are turned face down again and the next player proceeds. The player that collects more pairs wins the game. 

\begin{figure}
\centering
\includegraphics{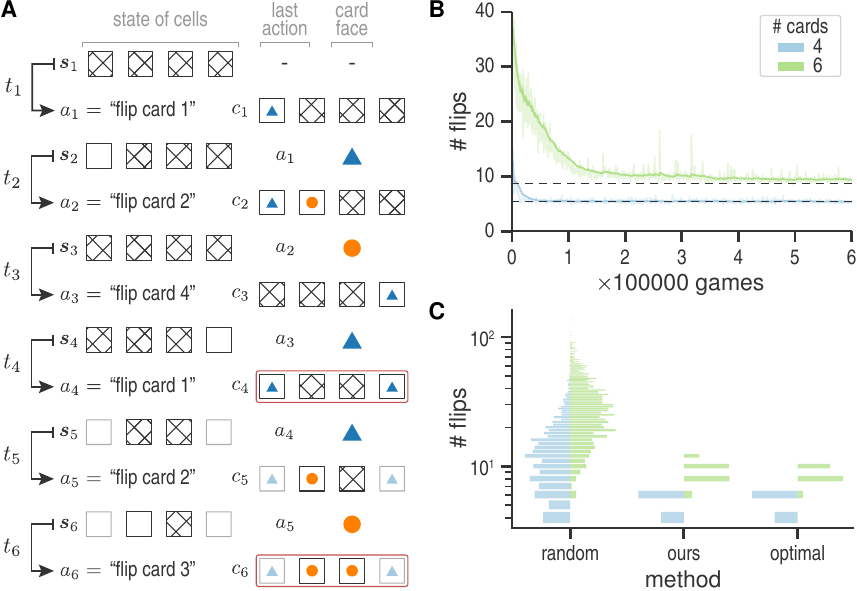}
\caption{An SNN learns to play the game \emph{Concentration} from rewards.
\textbf{(A)} Example game moves for a Concentration game with four cards played as a solitaire. The objective of the game is to turn over pairs of matching cards with as few card flips as possible.
Shown is---for each time step $t_i$---the agent's observation $\vm{s}_i$, the action $a_i$ taken by the agent, and the resulting card configuration $c_i$ after taking action $a_i$. The agent's observation $\vm{s}$ contains the state of the cells (face down card, illustrated by square with diamond pattern; face up card, square; or empty, grayed out square), the previous action taken by the agent, and the face of the card the agent had flipped in the previous time step. At time $t_1$ all cards are face down.
Flipping card \num{1} (action $a_1$) results in configuration $c_1$ (i.e., card \num{1}, showing a blue triangle, is face up).
Flipping card \num{2} in time step $t_2$ reveals an orange disc (configuration $c_2$). The two cards do not match, hence they are turned face down again in the next time step $t_3$.
In time step $t_3$, card \num{4}, that shows a blue triangle, is flipped. By recalling that the matching card is card \num{1}, the agent flips card \num{1} in $t_4$. Matching face up cards are then removed from the board. The game continues until the remaining two cards are removed. Board configurations at which the agent receives a reward are indicated by a red rectangle.
\textbf{(B)} Evolution of the number of card flips the agent takes to finish a game over the number of games during training. Shown is the evolution of the number of flips for a deck of four cards (blue) and six cards (green). The mean number of flips is shown as saturated solid line. Black dashed lines show the mean number of flips required when the agent has perfect memory and follows an optimal strategy (\num{5.33} for a deck of four cards and \num{8.65} for a deck of six cards).
\textbf{(C)} Histogram of the number of flips an agent takes to finish a game with four cards (blue) and six cards (green). Shown is the histogram for a random agent (left), our agent after training (middle), and an agent that has perfect memory and that follows an optimal strategy (right). Histograms are computed from \num{1000} games each and are scaled to the same width.
}
\label{fig:fig5}
\end{figure}

Here we consider a one-player solitaire version of the game (Fig.~\ref{fig:fig5}A). In this version of the game the objective is to find all matching pairs with as few card flips as possible. The cards are arranged on a one-dimensional grid of cells, each of which may be empty or may contain one card. The grid is just large enough to hold all of the $2n$ cards. Each card may either be face up or face down on any given time step (initially all cards are face down). The agent's available actions are to flip over any of the cards at a given time step. More precisely, the action space is an integer from $\{1, 2, \dots, 2n\}$. Whenever two cards are face up but do not match, they automatically turn face down in the next time step. Whenever two cards match they are removed from the grid and the agent is rewarded. The agent receives a small penalty for each card flip. The game continues until all cards are removed.

Instead of using images, we define each card face to be a \num{10}-dimensional random continuous-valued vector.
The agent's observation vector $\vm{s}$ (see Fig.~\ref{fig:fig5}A) contains three components: (a) a one-hot vector for each cell that encodes the state of the cell (a cell can either be empty, containing a face down card, or a face up card), (b) a one-hot vector encoding the previous action taken by the agent (i.e, a one-hot vector encoding which grid position was flipped), and (c) a 10-dimensional real vector for the image of the card the agent had flipped in the previous time step (this is a zero vector if the card is face down after the flip or if the agent's action was to flip an empty cell). 
In contrast to the other considered tasks, where we sequentially presented some facts followed by a query to which the network should respond with an output, here instead, in each time step, the network had to figure out by itself when to store and recall information, given only the current observation vector.

The performance was evaluated in terms of the number of flips performed until all matching pairs had been removed from the grid. Agents were trained with \gls{PPO} \cite{schulman2017proximal} in actor-critic style using \num{64} asynchronous vectorized environments (see Section \emph{Model and training details} in the Supplementary for details to the model and the training setup). We evaluated our model on a deck of four cards and a deck of six cards. The evolution of the number of card flips the agent takes to finish a game over the number of training steps is shown Fig.~\ref{fig:fig5}B. After training we evaluated the agents on \num{1000} games and recorded the number of card flips the agent takes to finish each game. Fig~\ref{fig:fig5}C shows the histogram of the number of card flips in this evaluation for a random agent, our agent, and an agent that has perfect memory and that follows an optimal strategy. If an agent has no memory at all, and plays by simply flipping cards at random, then the expected number of flips the agent takes to finish a game with $n$ pairs of cards is $(2n)^2$. For an optimal agent the length lies between $2n$ and $4n - 2$ where the expected number of flips is $(6 - 4\ln{2})n + 7/8 - 2\ln{2}$ as $n \to \infty$ \cite{velleman2013expect}. For the four card-game, the SNN reached an optimal performance (mean number of flips: \num{5.33}; optimal: \num{5.33}). An average of \num{8} flips were achieved within \num{8235} games. The six-card game was harder to train. Still, the final network's performance was again close to optimal (see Fig.~\ref{fig:fig5}C). Re-drawing the random vectors at the beginning of each game (i.e., using a new deck of cards in each game) marginally decreased the performance (mean number of flips was \num{5.35} for the four-card game, and \num{10.88} for the six-card game, where the optimum is \num{8.65} flips).



\section{Discussion}
We have presented a novel SNN model that integrates Hebbian plasticity in its network dynamics. We found that this memory-enrichment renders SNNs surprisingly flexible.

While only local plasticity is needed during inference, our model was trained with \gls{BPTT}. Since \gls{BPTT} is biologically implausible, we cannot claim that our network is a model for how the functionality could be learned by an organism. 
Instead, our results provide an existence proof for powerful memory-enhanced SNNs. One can speculate that brain networks were shaped by evolution to make use of Hebbian plasticity processes. In this sense, \gls{BPTT} can be seen as a replacement for evolutionary processes. For example, the brain might have evolved networks for one-shot learning (Fig.~\ref{fig:fig3}) that are particularly tuned to behaviorally relevant stimuli. In addition to evolutionary optimization, local approximations of \gls{BPTT} such as the recently introduced e-prop algorithm \cite{bellec2020solution} could then further shape the evolved circuits for specific functionality.

The integration of synaptic plasticity for inference in artificial neural networks was used in \cite{hinton1987using,schmidhuber1992learning} and adopted recently \cite{ba2016using,munkhdalai2018metalearning}. In the latter, input representations were bound to labels with Hebbian plasticity. 
Memory-augmented neural networks use explicit memory modules which are a differentiable version of a digital memory \cite{weston2015towards,sukhbaatar2015end,graves2016hybrid,miller2016key,xiong2016dynamic,chandar2016hierarchical,gulcehre2018dynamic}. Our model utilizes biological Hebbian plasticity instead. In \cite{miconi2018differentiable}, training of networks with synaptic plasticity was explored, but there the parameters of the plasticity rule were optimized instead of the surrounding control networks.

The inclusion of Hebbian plasticity can be viewed as the introduction of another long time constant in the network dynamics. Previous work has shown that longer time constants can significantly improve the temporal computing capabilities of SNNs. In this direction, short-term synaptic plasticity \cite{mongillo2008synaptic,maass2002real} and neuronal adaptation \cite{salaj2021spike} have been exploited. One-shot learning of SNNs has been studied in \cite{scherr2020one}. Instead of Hebbian plasticity, this model relied on a more elaborate three-factor learning rule. Another SNN model, the Spiking RelNet, was tested on the bAbI task set \cite{plank2021long}. We have compared this model to ours in Table \ref{tab:tab1}. The architecture of this model is quite different from our proposal. As a spiking implementation of relational networks, it is rather complex and makes heavy use of weight sharing. Both these models perform well on similar tasks of the bAbI task set, but interestingly, there are some differences. For example, our model solves task \num{2} "two supporting facts" on which the Spiking RelNet fails. This might be due to the possibility of multiple memory accesses through the feedback loop in our model. On the other hand, our model fails at task \num{17} "positional reasoning" which is solved by the Spiking RelNet. We suspect that this is due to the more complex network structure of the Spiking RelNet.
To the best of our knowledge, no previous spiking (or artificial) neural network model is performing well on both, one-shot learning and bAbI tasks, as well as on the other tasks we presented.

Hebbian plasticity is spatially and temporally local, i.e., the synaptic weight change depends only on a filtered version of the pre- and post-synaptic spikes and on the current weight value. This is a very desirable feature of any plasticity rule, as it can easily be implemented both in biological synaptic connections, and in neuromorphic hardware. In fact, current neuromorphic designs support this type of plasticity \cite{davies2018loihi}. Hence, our results indicate that Hebbian plasticity can serve as a fundamental building block in cognitive architectures based on energy-efficient neuromorphic hardware.

\section{Methods}

\subsection{Neuron Model}
\label{sec:methods-neuron_model}
Neurons in our model are standard \gls{LIF} neurons modeled in discrete time steps $\Delta t$. The membrane potential $V_j$ of a neuron $j$ at time $t$ is given by
\begin{equation}\label{eq:methods-membrane_potential}
    V_j(t + \Delta t) = \alpha V_j(t) + (1 - \alpha) I_j(t) - \vartheta z_j(t),
\end{equation}
where $\alpha = \exp(-\frac{\Delta t}{\tau_{\mathrm{m}}})$ defines the membrane potential decay per time step. The total synaptic input current $I_j$ is defined as the weighted sum of spikes from pre-synaptic neurons
\begin{equation}\label{eq:methods-syn_current}
    I_j(t) = \sum_i W_{ji} z_i(t),
\end{equation}
where the sum goes over all pre-synaptic neurons $i$ and where $W_{ji}$ is the weight of the corresponding synapse. A neuron $j$ spikes as soon as its membrane potential $V_j$ exceeds a firing threshold $\vartheta$. The spike train of a neuron $j$ is modeled as binary sequence $z_j \in \{0, 1\}$. After having fired a spike ($z_j(t) = 1$), the neurons membrane potential is reset by subtracting the threshold value $\vartheta$ (last term in \eqref{eq:methods-membrane_potential}) and the neuron enters an absolute refractory period $\Delta_{\mathrm{abs}}$ where it cannot spike again (for parameter values, see Table~\ref{tab:methods-neuron_parameters}).

On the basis of the above formalism, it is apparent that only the spikes are communicated to other neurons. Hence, we interpret $z_j$ as the output of a neuron $j$ and $V_j$ as its hidden state. For simplicity, in the following, we will not state the hidden dynamics of the neurons in the model, but only the synaptic input current and the output spike train.

\subsection{Model}
\label{sec:methods-model}
Our model takes a sequence of inputs $\langle x_1, \dots, x_M \rangle$. Each input $x_m$ can either be a fact or a query to which the network should respond with an output $\hat{\vm{o}}$.

\subsubsection{Input encoder}
Let $\langle x_1, \dots , x_M \rangle$ be the given input sequence. Each $x_m$ is converted into a spike train of duration $\tau_{\mathrm{sim}} = \SI{100}{ms}$ by using an input encoder (IE). We do not restrict the type of input encoder. It has to be chosen for the task at hand. In the simplest case we use a single dense layer $\mathcal{E}$ consisting of $d$ \gls{LIF} neurons.
Depending on whether the input $x_m$ represents some fact that might be useful to store, or a query to which the network should respond with an output, we denote the resulting spike train as $z_i^{\mathrm{s,enc}}$ and $z_i^{\mathrm{r,enc}}$, respectively.

\subsubsection{Memory induction}
Consider an input from which some aspect should be stored in memory. Here, the spike train encoding that input, that is $z_i^{\mathrm{s,enc}}$, is fed to two single-layer networks $\mathcal{K}$ and $\mathcal{V}$ comprising of $l$ LIF neurons each. The input current to a neuron $j$ in layer $\mathcal{K}$---we call it the key-layer---is given by
\begin{equation}
    I^{\mathrm{key}}_j(t) = \sum_i W^\mathrm{s,key}_{ji} z_i^{\mathrm{s,enc}}(t),
\end{equation}
where $W^{\mathrm{s,key}}$ is a synaptic-weight matrix of size $l \times d$. We denote the spike train of a neuron $j$ in the key-layer as $z_j^{\mathrm{key}}$. The input current to a neuron $k$ in the value-layer, layer $\mathcal{V}$, is given by
\begin{equation}
        I^{\mathrm{value}}_k(t) = \sum_i W^\mathrm{s,value}_{ki} z_i^{\mathrm{s,enc}}(t) + c \sum_j W^{\mathrm{assoc}}_{kj}(t) z_j^{\mathrm{key}}(t),
\end{equation}
where $W^{\mathrm{s,value}}$ is a synaptic-weight matrix of size $l \times d$, $c = 0.2$ is a constant, and where $W^{\mathrm{assoc}}$ are association synapses represented as a square matrix of size $l \times l$. The first term represents the contribution of the spikes from the input encoder to this current, and the second term represent the contribution of the spikes that travel from the key-layer via $W^{\mathrm{assoc}}$ to a neuron in layer $\mathcal{V}$. We denote the spike train of a neuron $k$ in this layer as $z_k^{\mathrm{value}}$.

Synapses $W^{\mathrm{assoc}}$ are subject to Hebbian plasticity. An association between the activity in the key- and value-layer neurons is established via weight changes given by:
\begin{equation}
    \label{eq:methods-hebb_update}
    \Delta W_{kj}^{\mathrm{assoc}}(t) = \gamma_+ (w^{\mathrm{max}} - W_{kj}^{\mathrm{assoc}}(t))\,\kappa_k^{\mathrm{value}}(t) \kappa_j^{\mathrm{key}}(t) - \gamma_- W_{kj}^{\mathrm{assoc}}(t) \kappa_j^{\mathrm{key}}(t)^2,
\end{equation}
where $\gamma_+ > 0$, $\gamma_- > 0$, $w^{\mathrm{max}}$ are constants, and where $\kappa_j^{\mathrm{key}}$ and $\kappa_k^{\mathrm{value}}$ are exponential activity traces of $z_j^{\mathrm{key}}$ and $z_k^{\mathrm{value}}$, respectively (for parameter values, see Table~\ref{tab:methods-neuron_parameters}). Traces are updated by an amount $1 - \exp(-\Delta t/\tau_{\mathrm{trace}})$ at the moment of spike arrival and decay exponentially with time constant $\tau_{\mathrm{trace}}$ in the absence of spikes.

The first term in \eqref{eq:methods-hebb_update} implements a soft upper bound $w^{\mathrm{max}}$ on the weights. The Hebbian term $\kappa_k^{\mathrm{value}}(t)\kappa_j^{\mathrm{key}}(t)$ strengthens connections between co-active neurons in the key- and value-layers. Finally, the last term generally weakens connections from the currently active key-neurons. Since the Hebbian component strengthens connections to active value-neurons, this emphasizes the current association and de-emphasizes old ones. This update is similar to Oja's rule \cite{oja1982simplified}, but note that the quadratic term acts on the pre-synaptic neuron. Association synapses are then updated according to $W^{\mathrm{assoc}}(t + \Delta t) = W^{\mathrm{assoc}}(t) + \Delta W^{\mathrm{assoc}}(t)$.

\subsubsection{Memory recall}
Now consider an input that should trigger a memory recall. In this case, we feed the encoded input $z_i^{\mathrm{r,enc}}$ and, via delayed feedback connections, the activity in the value layer $z_k^{\mathrm{value}}$ to the key-layer $\mathcal{K}$. The input current to a neuron $j$ in this layer is then given by
\begin{equation}
    I^{\mathrm{key}}_j(t) = \sum_{i \le d} W^\mathrm{r,key}_{ji} z_i^{\mathrm{r,enc}}(t) + \sum_{d \le k \le d+l} W^\mathrm{r,key}_{jk} z_{k-d}^{\mathrm{value}}(t-d_{\mathrm{feedback}}),
\end{equation}
where $W^{\mathrm{r,key}}$ is a synaptic-weight matrix of size $l \times d + l$, and where $d_{\mathrm{feedback}}$ is the synaptic delay in the feedback connections. As before, we denote the spike train of a neuron $j$ in this layer as $z_j^{\mathrm{key}}$. This activity activates some neurons in the value-layer through synapses $W^{\mathrm{assoc}}$, thus potentially recalling some information that has been stores previously. The synaptic current to a neuron $k$ in the value-layer is given by
\begin{equation}
    I^{\mathrm{value}}_{k}(t) = \sum_j W_{kj}^{\mathrm{assoc}}(t) z_j^{\mathrm{key}}(t).
\end{equation}
As before, we denote the spike train of a neuron $k$ in this layer as $z_k^{\mathrm{value}}$.

\subsubsection{Generating the final prediction}
The architecture of the output module depends on the task at hand. In the simplest case, the network output was determined by taking the sum of $z_k^{\mathrm{value}}$ over the last $\tau_{\mathrm{read}}$ time steps and passing the result through a final weight matrix $W_{\mathrm{out}}$:
\begin{equation}
    \hat{o}_j = \sum_k W^{\mathrm{out}}_{jk} \sum_{t^{\prime} = 0}^{\tau_{\mathrm{read}}} z_k^{\mathrm{value}}(M\tau_{\mathrm{sim}} - t^\prime).
\end{equation}
The weights $W^{\mathrm{s,key}}$, $W^{\mathrm{s,value}}$, $W^{\mathrm{r, key}}$, and $W^{\mathrm{out}}$ are learned during training by minimizing the cross-entropy loss between the softmax of $\hat{\vm{o}}$ and the target output $\vm{o}$ using the Adam optimizer \cite{kingma2014adam}. The gradient is backpropagated through spikes by replacing the non-existent derivative of the membrane potential at the time of a spike by a pseudo-derivative, as done in \cite{bellec2018long}, that smoothly increases from 0 to 1, and then decays back to 0:
\begin{equation}
    \frac{dz_j(t)}{dv_j(t)} \coloneqq \beta\,\mathrm{max}\{0, 1 - |v_j(t)|\}
\end{equation}
where $v_j(t) = \frac{V_j(t) - \vartheta}{\vartheta}$ is the normalized membrane potential of neuron $j$ and $\beta \le 1$ is a dampening factor which was set to $1$. Association synapses $W^{\mathrm{assoc}}$ were represented by a square matrix which was initialized for each input sequence $\langle x_m \rangle$ with all its values set to zeros.

\begin{table}
\caption{Neuron and plasticity parameters.}
\centering
\sffamily\scriptsize\selectfont
\begin{tabular}{@{\extracolsep{\fill}}
l
l
S[table-alignment=right]@{}}
\toprule
$\vartheta$ & Firing threshold & \num{0.1} \\
$\Delta_{\mathrm{abs}}$ & Refractory period & \SI{3}{ms} \\
$\tau_{\mathrm{m}}$ & Membrane time constant & \SI{20}{ms} \\
$\tau_{\mathrm{trace}}$ & Time constant of synaptic trace & \SI{20}{ms} \\
$w^{\mathrm{max}}$ & Soft maximum of Hebbian weights & \num{1.0} \\
$\gamma_{+}$ & Write factor of Hebbian update rule & \num{0.3} \\
$\gamma_{-}$ & Forget factor of Hebbian update rule & \num{0.3} \\
\bottomrule
\end{tabular}
\label{tab:methods-neuron_parameters}
\end{table}

\subsection*{Data Availability}
The Omniglot data set \cite{lake2015human} is freely available at \href{https://github.com/brendenlake/omniglot}{\nolinkurl{https://github.com/brendenlake/omniglot}}. The FSDD data set \cite{fsd2016} is freely available at \href{https://github.com/Jakobovski/free-spoken-digit-dataset}{\nolinkurl{https://github.com/Jakobovski/free-spoken-digit-dataset}}. The MNIST data set \cite{lecun2010mnist} is freely available at \href{http://yann.lecun.com/exdb/mnist}{\nolinkurl{http://yann.lecun.com/exdb/mnist}}. The bAbI data set \cite{weston2015towards} is freely available at \href{https://research.fb.com/downloads/babi}{\nolinkurl{https://research.fb.com/downloads/babi}}.

\subsection*{Code Availability}
The models were implemented in PyTorch \cite{paszke2019pytorch} and the code is available at \href{https://github.com/IGITUGraz/MemoryDependentComputation}{\nolinkurl{https://github.com/IGITUGraz/MemoryDependentComputation}}.

\subsection*{ACKNOWLEDGMENTS}
This work was supported by the CHIST-ERA grant CHIST-ERA-18-ACAI-004, by the Austrian Science Fund (FWF) project number I 4670-N (project SMALL), and by the "University SAL Labs" initiative of Silicon Austria Labs (SAL). We thank Wolfgang Maass and Arjun Rao for initial discussions.


\printbibliography

\clearpage
\appendix
\setcounter{equation}{0}
\setcounter{figure}{0}
\setcounter{table}{0}
\renewcommand{\thefigure}{S\arabic{figure}}
\renewcommand{\thetable}{S\arabic{table}}

\section*{Supplementary}

\subsection*{Model and training details}
Here we give details to our models, and to the encoding and representation of inputs used in our models. The models were simulated with a time step $\Delta t$ of \SI{1}{ms}. We used the Euler method for the integration of the neuron dynamics. Every neuron state, that is, membrane potential, remaining refractory period, etc. are represented by fixed size state vectors that are updated at each time step of the simulation.

\subsubsection*{Spike rate regularization}
\label{sec:supp-rate_regularization}
To achieve biologically plausible neuron and network activity, we used an
$L_2$ activity regularizer, which encourages solutions with sparse spiking. For any particular layer with $N$ neurons, the spike rate regularization loss is calculated according to
\begin{equation}
    L_{\rho} = \lambda_{\rho} \frac{1}{N}\sum_{i=1}^N (\rho_0 - \rho_i)^2,
\label{eq:supp-rate_regularization}
\end{equation}
where $\lambda_{\rho}$ is the regularization factor, $\rho_0 = 0$ is the target firing rate, and $\rho_i$ is the average firing rate of neuron $i$. The average is across the entire simulation duration of a given layer and the entire batch.

\subsubsection*{Converting pre-trained CNNs}
\label{sec:supp-cnn_conversion}
We considered the pre-trained \gls{CNN} parameters as the synaptic weights while computing the weighted sum-of-spikes from pre-synaptic neurons via convolution operations. We replaced the post-convolution ReLU activations with \gls{IF} neuron dynamics which implements $V_j(t + \Delta t) = V_j(t) + I_j(t) - \vartheta z_j(t)$ without membrane potential leaks within the \gls{CNN} layers, as opposed to the \gls{LIF} neurons which are used in our models otherwise. Finally, we determine the layer-wise \gls{IF} neuron firing thresholds within the \gls{CNN} using the ANN-to-SNN conversion by threshold-balancing algorithm \cite{diehl2015fast,sengupta2019going}. The algorithm obtains a firing threshold for each \gls{CNN} layer sequentially. We initialize all four layer-wise \gls{CNN} firing thresholds with zeros. We treat the model inputs (i.e., grayscale Omniglot/MNIST pixel intensities or \gls{MFCC} values) as a constant current applied for \SI{100}{\ms} to \gls{LIF} neurons, to produce input spikes to this \gls{CNN}. We used training set samples converted into input spikes (all training samples for MNIST and \gls{MFCC} encoders, and \num{12800} randomly drawn images for Omniglot which we found to be sufficient to obtain stable threshold estimates) and performed an initial forward pass from the first layer of the \gls{CNN}. While doing so, we record the value of the maximum input current (incoming weighted sum-of-spikes) observed across all used training samples at any time step during the \SI{100}{\ms}, and set the firing threshold of the first \gls{CNN} layer to have this value at the end. After the firing threshold of the first layer is set, we perform a forward pass of these spiking inputs starting from the first \gls{CNN} layer once again using the set firing threshold, and then determine the firing threshold for the second layer in a similar way. We continue this process sequentially for all four layers of the \gls{CNN} to determine layer-wise firing thresholds.

Following ANN-to-SNN conversion, spiking \gls{CNN} synaptic weights were fine-tuned during end-to-end training of the models. At this stage, in cross-modal associations experiments, we also introduced additional (non-zero) bias terms for all convolution kernels in the \glspl{CNN}. Note that this allows us to also fine-tune the conversion-based preset layer-wise firing thresholds individually for each \gls{CNN} convolution kernel. In one-shot learning experiments, however, we did not (find it beneficial to) introduce bias terms to fine-tune the preset layer-wise \gls{CNN} firing thresholds.

\subsubsection*{Details to: Memorizing associations}
We used two dense layers as input encoder in this task. Each layer consisted of \num{80} \gls{LIF} neurons. One layer was used to encode each of the random input vectors and another layer was used to encode the integer labels. We trained models using different values for N (ranging from \num{2} to \num{55}). We tuned the hyper-parameters using a validation set which contained \num{1000} sequences. The models were tested on \num{2000} sequences. We generated a data set for every iteration of \gls{BPTT}. We trained for \num{4250} iterations using mini batches of size \num{512}. The models were trained with Adam \cite{kingma2014adam} using a learning rate of $\mu = 0.003$, that was reduced by \SI{15}{\percent} every \num{340} iterations. The output of the network was produced by passing the number of spikes over the last \SI{30}{ms} of $\vm{z}^\mathrm{value}$ through a final output layer with weights $W^\mathrm{out}$ and a softmax. The error was computed when the network produced its prediction after the query and gradients were propagated through all time steps of the computation. Weights were initialized using Glorot uniform initialization \cite{glorot2010understanding} with a gain of $\sqrt{2}$. The feedback delay $d_{\mathrm{feedback}}$  was set to \SI{1}{ms}. Starting from iteration \num{0}, we applied $L_2$ activity regularization to all spiking neurons of the model (see Section \emph{Spike rate regularization} for details). The regularization factor $\lambda_{\rho}$ was set to \num{1e-5}. Gradients with an $L_2$-norm larger than \num{40.0} were normalized to have norm \num{40.0}.

\subsubsection*{Details to: One-shot Learning}
We used a \gls{CNN} as input encoder for the Omniglot images.
The \gls{CNN} consisted of four weight layers and had the following structure: $4$ $\times$ (Conv2D $\rightarrow$ Activation $\rightarrow$ MaxPooling) $\rightarrow$ Flatten. Each block comprises a \num{64}-filter convolution with a kernel size of $3 \times 3$ (stride = \num{1}, zero-padding = \num{1} for each dimension) without bias terms. We used a $2 \times 2$ pool size in the max pooling layers. 
When applied to the $28 \times 28$ Omniglot images this architecture results in a \num{64}-dimensional output space. 
This \gls{CNN} was pre-trained using the prototypical loss \cite{snell2017prototypical} with a ReLU function as the activation after convolutions (i.e., as a conventional \gls{ANN}).
We converted the pre-trained \gls{CNN} into a spiking \gls{CNN} that uses \gls{IF} neuron dynamics as post-convolution activations, and used a threshold-balancing algorithm to set layer-wise firing thresholds \cite{diehl2015fast,sengupta2019going} (detailed in Section \emph{Converting pre-trained CNNs}).
The converted spiking \gls{CNN} synaptic weights were later fine-tuned during end-to-end training.
We treated the grayscale values of an Omniglot image as a constant input current, applied for \SI{100}{\ms} to \num{784} \gls{LIF} neurons, to produce input spikes to this \gls{CNN}.
The final layer of the \gls{CNN} consisted of \num{64} \gls{LIF} neurons. 
A single dense layer consisting also of \num{64} \gls{LIF} neurons was used to encode the integer labels as spike trains with a duration of \SI{100}{\ms} per label.

We tuned the hyper-parameters using a held-out validation set of \num{172} characters (\num{688} classes with rotations). We trained for \num{200} epochs with \num{200} iterations per epoch using mini batches of size \num{256}. The models were trained with Adam \cite{kingma2014adam} using a learning rate of $\mu = 0.001$, that was reduced by \SI{15}{\percent} every \num{20} epochs. The output of the network was produced by taking the sum of spikes over the last \SI{30}{ms} of $\vm{z}^\mathrm{value}$, and by passing that value through a final output layer with weights $W^\mathrm{out}$ and a softmax. The error was computed when the network produced its prediction after the query and gradients were propagated through all time steps of the computation. Weights were initialized using Glorot uniform initialization \cite{glorot2010understanding} with a gain of $\sqrt{2}$. The feedback delay $d_{\mathrm{feedback}}$ was set to \SI{1}{ms}. We applied $L_2$ activity regularization after an initial training period of one epoch to all spiking neurons of the model (see Section \emph{Spike rate regularization} for details). The regularization factor $\lambda_{\rho}$ was set to \num{1e-6}. Gradients with an $L_2$-norm larger than \num{40.0} were normalized to have norm \num{40.0}.

\subsubsection*{Details to: Cross-modal associations}
We used version 1.0.10 of the FSDD data set available on GitHub, and the MNIST data set from the torchvision data set API (we kept the default train-test split of these data sets). We used one \gls{CNN} to encode the audio input and one \gls{CNN} to encode the MNIST images. We extracted \num{20} \glspl{MFCC} from the audio signal before feeding it to the convolutional encoder network (we computed \num{21} \glspl{MFCC}, but we discarded the constant offset coefficient, that is, the \num{0}th coefficient; \glspl{MFCC} were zero padded to \num{30} time samples and scaled such that each coefficient dimension had zero mean and unit variance).
The \glspl{CNN} consisted of four weight layers and had the following structure: $4$ $\times$ (Conv2D $\rightarrow$ Activation $\rightarrow$ MaxPooling) $\rightarrow$ Flatten. Each block comprises a \num{64}-filter convolution with a kernel size of $3 \times 3$ (stride = \num{1}, zero-padding = \num{1} for each dimension) without bias terms. We used a $2 \times 2$ pool size in the max pooling layers. 
When applied to the $30 \times 20$ \gls{MFCC} features, to the $28 \times 28$ MNIST images, this architecture results in a \num{64}-dimensional output space. 
These \glspl{CNN} were respectively pre-trained on FSDD/MNIST \num{10}-class digit classification tasks with ReLU functions used as activations after convolutions (i.e., as a conventional \gls{ANN}), and additional $64 \times 10$ dimensional linear classification layers at the output.
Following pre-training we discarded the final classification layers.
We converted the pre-trained \glspl{CNN} into spiking \glspl{CNN} that uses \gls{IF} neuron dynamics as post-convolution activations, and used a threshold-balancing algorithm to set layer-wise firing thresholds \cite{diehl2015fast,sengupta2019going} (detailed in Section \emph{Converting pre-trained CNNs}).
The converted spiking \gls{CNN} synaptic weights were later fine-tuned during end-to-end training.
Input spikes to the \glspl{CNN} were produced by applying the grayscale values of the MNIST images and \gls{MFCC} features to \num{784} and \num{600} \gls{LIF} neurons, respectively (we treated these values as constant input current over \SI{100}{\ms} per input).
The final layers of both \glspl{CNN} consisted of \num{64} \gls{LIF} neurons encoding the input stimuli.

We tuned the hyper-parameters using a validation set which contained \num{1000} sequences. To encourage the model to learn to associate a specific instance of the FSDD data set to a specific instance of the MNIST data set, we generated training examples as follow: We randomly choose three digits between \num{0} and \num{9}. We then generated three tuples each containing a randomly drawn instance of these digits as audio file from the FSDD data set, and a randomly drawn instance of these digits as image from the MNIST data set. The query was another randomly drawn sample from one of the FSDD classes in the sequence. Test examples were generated as described in the main text. 
The image reconstruction was produced by a two-layer fully-connected network with \num{256} and \num{784} \gls{LIF} neurons, respectively. The image was reconstructed by taking the sum of spikes of each neuron in the last layer, scaling the result by $\frac{1}{15}$, and reshaping it to $28 \times 28$. 
We trained our model for \num{4680} iterations using mini batches of size \num{256}. The models were trained with Adam \cite{kingma2014adam} using a learning rate of $\mu = 0.001$, that was reduced by \SI{15}{\percent} every \num{780} iterations. During training, all network weights are jointly learned by minimizing the \gls{MSE} between the output of the model and the target MNIST image. The error was computed when the network produced its output after the query and gradients were propagated through all time steps of the computation. Weights were initialized using Glorot uniform initialization \cite{glorot2010understanding} with a gain of $\sqrt{2}$. The feedback delay $d_{\mathrm{feedback}}$ was set to \SI{1}{ms}. Starting from iteration \num{0}, we applied $L_2$ activity regularization to all spiking neurons of the model (see Section \emph{Spike rate regularization} for details). The regularization factor $\lambda_{\rho}$ was set to \num{1e-7}. Gradients with an $L_2$-norm larger than \num{40.0} were normalized to have norm \num{40.0}.

\subsubsection*{Details to: Question Answering}
We used version 1.2 of the bAbI data set (we kept the default train-test split of the data set). We used a learned encoding (LE) for sentences as proposed in \cite{henaff2017tracking}. The encoding is given by $\vm{e}_m = \sum_j \vm{f}_j \circ A\vm{w}_{m,j}$, where $\vm{w}_{m,j}$ is a word of a sentence $x_m = \{\vm{w}_{m,1}, \vm{w}_{m,2}, \dots, \vm{w}_{m,J}\}$ and where $A$ is the embedding matrix and $\circ$ denotes the Hadamard product. The vectors $\vm{f}_j$ were constant across time steps and were trained jointly with the other parameters of our model. We treated $\vm{e}_m$ as constant input current, applied for \SI{100}{ms} to a layer of \num{80} \gls{LIF} neurons, to produces input spikes to our model. We tuned the hyper-parameters on a held-out validation set which was \SI{10}{\percent} of the training set. The networks were trained for \num{200} epochs on \num{10000} examples per task with a batch size of \num{256}. The models were trained with Adam \cite{kingma2014adam} using a learning rate of $\mu = 0.003$, that was reduced by \SI{15}{\percent} every \num{20} epochs. The output of the network was produced by taking the sum of spikes over the last \SI{30}{ms} of $\vm{z}^\mathrm{value}$, and by passing that value through a final output layer with weights $W^\mathrm{out}$ and a softmax. The error was computed when the network produced its prediction after the query and gradients were propagated through all time steps of the computation. Weights were initialized using Glorot uniform initialization \cite{glorot2010understanding} with a gain of $\sqrt{2}$. We applied $L_2$ activity regularization after an initial training period of one epoch to all spiking neurons of the model (see Section \emph{Spike rate regularization} for details). The regularization factor $\lambda_{\rho}$ was set to \num{1e-5}. Gradients with an $L_2$-norm larger than \num{40.0} were normalized to have norm \num{40.0}. Since the number of sentences and the number of words per sentence varied within and between tasks, a null symbol was used to pad them to a fixed size. The embedding of the null symbol was constraint to be zero.

\subsubsection*{Details to: Reinforcement Learning}
We used a dense layer consisting of \num{80} \gls{LIF} neurons as input encoder in this task. Agents were trained with \gls{PPO} \cite{schulman2017proximal} in actor-critic style using \num{64} asynchronous vectorized environments. We trained for \num{4000} iterations. In each iteration, we performed \num{10} steps (\num{100} steps for 6-card games) in each environment. The surrogate loss was then updated over \num{4} epochs and \num{16} training mini batches using Adam \cite{kingma2014adam} with a learning rate of $\mu = 0.0003$. We used a value function coefficient of \num{0.1}, an entropy coefficient of \num{0.01}, and a discount factor of \num{0.9}. The \gls{PPO} clip parameter was set to \num{0.2}. The agent received a reward of \num{25} for finding a matching pair and a small penalty of \num{0.5} for each card flip. The feedback delay $d_{\mathrm{feedback}}$ was set to \SI{1}{ms}. Gradients with an $L_2$-norm larger than \num{0.5} were normalized to have norm \num{0.5}. Actions were produced by taking the sum of spikes over the last \SI{30}{ms} of $\vm{z}^\mathrm{value}$, and by passing that value through a three-layer fully-connected network with \num{100}, \num{100}, and \num{4} (\num{6} for 6-card games) neurons, respectively. We used hyperbolic tangent (tanh) activation functions except for the last layer. The critic was also a three-layer fully-connected network comprising of \num{100}, \num{100}, and \num{1} neurons, respectively. Again, we used tanh activation functions except for the last layer. The critic network received as input, in addition to the sum of spikes over the last \SI{30}{ms} of $\vm{z}^\mathrm{value}$, also the agent's observations. Actor and critic networks were initialized using orthogonal initialization \cite{saxe14exactsolutions} with a gain of $\sqrt{2}$. Biases in these networks were zero initialized. Other weights in our model were initialized using Glorot uniform initialization \cite{glorot2010understanding} with a gain of $\sqrt{2}$.

\begin{table}
\caption{Sample stories and questions from the bAbI data set \cite{weston2015towards}.}
\centering
\scriptsize\sffamily\selectfont
\begin{tabular}{@{}lll@{}}
\textbf{Task 1: Single Supporting Fact}  &
\textbf{Task 2: Two Supporting Facts}    &
\textbf{Task 3: Three Supporting Facts}    \\
\cmidrule(r){1-1} \cmidrule(lr){2-2} \cmidrule(lr){3-3}
Mary went to the bathroom.            & John is in the playground.       & John picked up the apple.                  \\
John moved to the hallway.            & John picked up the football.     & John went to the office.                   \\
Mary travelled to the office.         & Bob went to the kitchen.         & John went to the kitchen.                  \\
Q: Where is Mary?                     & Q: Where is the football?        & John dropped the apple.                    \\
A: office                             & A: playground                    & Q: Where was the apple before the kitchen? \\
                                      &                                  & A: office                                  \\
&&\\
\textbf{Task 4: Two Argument Relations}  &
\textbf{Task 5: Three Argument Relations}    &
\textbf{Task 6: Yes/No Questions}    \\
\cmidrule(r){1-1} \cmidrule(lr){2-2} \cmidrule(lr){3-3}
The office is north of the bedroom.   & Mary gave the cake to Fred.      & John moved to the playground. \\
The bedroom is north of the bathroom. & Fred gave the cake to Bill.      & Daniel went to the bathroom.  \\
The kitchen is west of the garden.    & Jeff was given the milk by Bill. & John went back to the hallway.\\
Q: What is north of the bedroom?      & Q: Who gave the cake to Fred?    & Q: Is John in the playground? \\
A: office                             & A: Mary                          & A: no                         \\
&&\\
\textbf{Task 7: Counting}  &
\textbf{Task 8: Lists/Sets}    &
\textbf{Task 9: Simple Negation}    \\
\cmidrule(r){1-1} \cmidrule(lr){2-2} \cmidrule(lr){3-3}
Daniel picked up the football.         & Daniel picks up the football.    & Sandra travelled to the office. \\
Daniel dropped the football.           & Daniel drops the newspaper.      & Fred is no longer in the office.\\
Daniel got the milk.                   & Daniel picks up the milk.        & Q: Is Fred in the office?       \\
Daniel took the apple.                 & John took the apple.             & A: no                           \\
Q: How many objects is Daniel holding? & Q: What is Daniel holding?       & Q: Is Sandra in the office?     \\
A: two                                 & A: milk, football                & A: yes                          \\
&&\\
\textbf{Task 10: Indefinite Knowledge}  &
\textbf{Task 11: Basic Coreference}    &
\textbf{Task 12: Conjunction}    \\
\cmidrule(r){1-1} \cmidrule(lr){2-2} \cmidrule(lr){3-3}
John is either in the classroom or the garden. & Daniel was in the kitchen.  & Mary and Jeff went to the kitchen. \\
Sandra is in the garden.                           & Then he went to the studio. & Then Jeff went to the park.        \\
Q: Is John in the classroom?                       & Sandra was in the office.   & Q: Where is Mary?                  \\
A: maybe                                           & Q: Where is Daniel?         & A: kitchen                         \\
Q: Is John in the office?                          & A: studio                   & Q: Where is Jeff?                  \\
A: no                                              &                             & A: park                            \\
&&\\
\textbf{Task 13: Compound Coreference}  &
\textbf{Task 14: Time Reasoning}    &
\textbf{Task 15: Basic Deduction}    \\
\cmidrule(r){1-1} \cmidrule(lr){2-2} \cmidrule(lr){3-3}
Daniel and Sandra journeyed to the office. & In the afternoon Julie went to the park. & Sheep are afraid of wolves.    \\
Then they went to the garden.              & Yesterday Julie was at school.           & Cats are afraid of dogs.       \\
Sandra and John travelled to the kitchen.  & Julie went to the cinema this evening.   & Mice are afraid of cats.       \\
After that they moved to the hallway.      & Q: Where did Julie go after the park?    & Gertrude is a sheep.           \\
Q: Where is Daniel?                        & A: cinema                                & Q: What is Gertrude afraid of? \\
A: garden                                  & Q: Where was Julie before the park?      & A: wolves                      \\
                                           & A: school                                &                                \\
&&\\
\textbf{Task 16: Basic Induction}  &
\textbf{Task 17: Positional Reasoning}    &
\textbf{Task 18: Size Reasoning}    \\
\cmidrule(r){1-1} \cmidrule(lr){2-2} \cmidrule(lr){3-3}
Lily is a swan.        & The triangle is to the right of the blue square.      & The football fits in the suitcase.    \\
Lily is white.         & The red square is on top of the blue square.          & The suitcase fits in the cupboard.    \\
Bernhard is green.     & The red sphere is to the right of the blue square.    & The box is smaller than the football. \\
Greg is a swan.        & Q: Is the red sphere to the right of the blue square? & Q: Will the box fit in the suitcase?  \\
Q: What color is Greg? & A: yes                                                & A: yes                                \\
A: white               &                                                       &                                       \\
&&\\
\textbf{Task 19: Path Finding}  &
\textbf{Task 20: Agent's Motivations}    &
\\
\cmidrule(r){1-1} \cmidrule(lr){2-2}
The kitchen is north of the hallway.          & John is hungry.               & \\
The bathroom is west of the bedroom.          & John goes to the kitchen.     & \\
The den is east of the hallway.               & John grabbed the apple there. & \\
The office is south of the bedroom.           & Daniel is hungry.             & \\
Q: How do you go from the den to the kitchen? & Q: Where does Daniel go?      & \\
A: west, north                                & A: kitchen                    & \\
\end{tabular}
\label{table:sample-all-babi-tasks}
\end{table}

\end{document}